\UseRawInputEncoding
\documentclass[]{ceurart}

\usepackage[utf8]{inputenc}
\usepackage[T1]{fontenc}
\usepackage{amsmath}
\usepackage{graphicx}
\usepackage{multirow}
\usepackage{array}
\usepackage{booktabs}
\usepackage{adjustbox}
\usepackage{listings}
\begin{document}

\copyrightyear{2024}
\copyrightclause{Copyright for this paper by its authors.
  Use permitted under Creative Commons License Attribution 4.0
  International (CC BY 4.0).}

\conference{CLEF 2024: Conference and Labs of the Evaluation Forum, September 09–12, 2024, Grenoble, France}

\title{UIT-DarkCow team at ImageCLEFmedical Caption 2024: Diagnostic Captioning for Radiology Images Efficiency with Transformer Models}

% % \tnotetext[1]{You can use this document as the template for preparing your
% %   publication. We recommend using the latest version of the ceurart style.}

% \title[mode=sub]{Notebook for the <Lab name> Lab at CLEF 2024}

\author[1,3]{Quan Van Nguyen}[
orcid=0009-0000-3604-4679,
email=21521333@gm.uit.edu.vn
]

\address[1]{Faculty of Information Science and Engineering, University of Information Technology, Ho Chi Minh City, Vietnam}
\address[2]{Faculty of Computer Science, University of Information Technology, Ho Chi Minh City, Vietnam}
\address[3]{Vietnam National University, Ho Chi Minh City, Vietnam}

\author[1,3]{Huy Quang Pham}[
orcid=0009-0006-5815-4469,
email=21522163@gm.uit.edu.vn
]
\author[1,3]{Dan Quang Tran}[
orcid=0009-0003-8806-5289,
email=21521917@gm.uit.edu.vn
]

\author[1,3]{Thang Kien-Bao Nguyen}[
orcid=0009-0009-6456-4247,
email=21521432@gm.uit.edu.vn
]
\author[1,2]{Nhat-Hao Nguyen-Dang}[
orcid=0009-0007-6405-7603,
email=20520490@gm.uit.edu.vn
]

\author[1,3]{Bao-Thien Nguyen-Tat}[
orcid=0000-0002-4809-7126,
email=thienntb@uit.edu.vn
]
\cormark[1]
\cortext[1]{Corresponding author.}

\begin{abstract}
Purpose: This study focuses on the development of automated text generation from radiology images, termed diagnostic captioning, to assist medical professionals in reducing clinical errors and improving productivity. The aim is to provide tools that enhance report quality and efficiency, which can significantly impact both clinical practice and deep learning research in the biomedical field.\\
Methods: In our participation in the ImageCLEFmedical2024 Caption evaluation campaign, we explored caption prediction tasks using advanced Transformer-based models. We developed methods incorporating Transformer encoder-decoder and Query Transformer architectures. These models were trained and evaluated to generate diagnostic captions from radiology images.\\
Results: Experimental evaluations demonstrated the effectiveness of our models, with the VisionDiagnostor-BioBART model achieving the highest BERTScore of 0.6267. This performance contributed to our team, DarkCow, achieving third place on the leaderboard.\\
Conclusion: Our diagnostic captioning models show great promise in aiding medical professionals by generating high-quality reports efficiently. This approach can facilitate better data processing and performance optimization in medical imaging departments, ultimately benefiting healthcare delivery.
\end{abstract}

\begin{keywords}
ImageCLEF, Computer Vision, Diagnostic Captioning, Image Captioning, Image Understanding, Radiology Images, Transformer Models, Encoder-Decoder, Query Transformer
\end{keywords}

\maketitle

\section{Introduction}\label{sec1}
Machine learning, especially Deep Learning, is creating breakthroughs in many different fields, and its impact on biomedicine is remarkable. With the exponential growth of biomedical data, researchers are exploring its potential in biomedical engineering, advanced computing, imaging systems, and biomedical data mining algorithms based on machine learning \cite{lawson2021machine}. One important area is Diagnostic Captioning. Diagnostic Captioning is the process of automatically generating diagnostic text based on a set of medical images collected during a medical examination. It can assist less experienced physicians by minimizing clinical errors and helping experienced physicians generate diagnostic reports faster \cite{pavlopoulos2022diagnostic}.

ImageCLEF is an annual multimodal machine learning campaign, part of the Cross-Language Evaluation Forum (CLEF), which has been running since 2003. It encourages breakthroughs in research and development of processing systems. Advanced multimedia processing in computer vision, image analysis, classification and retrieval in a multilingual, multimodal context. This year, one of ImageCLEF's four main missions is ImageCLEFMedical, which includes a series of challenges from annotating images to creating synthetic images and answering questions. In ImageCLEF 2024 \cite{ImageCLEF2024}, we took part in the ImageCLEFmedical Caption task. As in previous years, this task comprised two subtasks: concept detection and caption prediction.

Concept detection aims to associate biomedical images with related medical concepts while captioning prediction focuses on automatically generating preliminary diagnostic reports that accurately describe medical conditions and structures and anatomy shown in images. Concept detection also supports diagnostic notes by identifying key concepts that should be included in the preliminary report. Additionally, it can be used to index medical images according to related concepts, facilitating more efficient organization and retrieval.

Captioning prediction, in other words, diagnostic captioning, remains a challenging research problem, designed to support the diagnostic process by providing a preliminary report rather than replacing the physicians and human factors involved \cite{pavlopoulos2022diagnostic}. It is designed as a tool to assist in generating an initial diagnostic report of a patient's condition, helping doctors focus on important areas of the image \cite{shin2016learning} and assisting them in making diagnoses. Guess more accurately quickly \cite{moschovis2022Medical}. This approach can increase the efficiency of experienced clinicians, allowing them to handle high volumes of daily medical examinations more quickly and efficiently. For less experienced clinicians, automated annotation can help reduce the likelihood of clinical errors\cite{pavlopoulos2019survey}.

\subsection{DarkCow Team Contributions}
In this paper, we presented the experiments and the systems that were submitted by our DarkCow team in this year's caption prediction task, which helped us secure third place on the leaderboard (see Table \ref{ranking}). Our new approaches build on the rapid development of deep learning techniques, especially the Transformer \cite{vaswani2017attention} encoder-decoder architecture and the Query Transformer \cite{wang2022anchor} for Large Language Model \cite{zhao2023survey}.
We exploited the power of Vision Transformer (ViT) to extract visual features from radiology images. To optimize the use of information, we also used VinVL to extract features of objects in the images. Our first approach is based on encoder-decoder architecture to generate image captions. In the second approach, we leveraged Query Transformer to help LLM understand images. We also conducted experiments with image pre-processing, caption length, and object features to analyze the impact of those aspects.

\begin{table}[ht]
\caption{Caption prediction task scores, rankings are based on BERTScore}
\label{ranking}
\centering
\begin{adjustbox}{max width=\textwidth}
\begin{tabular}{|l|c|c|c|c|c|c|c|c|c|c|c|}
\hline
\textbf{Team} & \textbf{ID} & \textbf{BERTScore} & \textbf{ROUGE} & \textbf{BLEU-1} & \textbf{BLEURT} & \textbf{METEOR} & \textbf{CIDEr} & \textbf{CLIPScore} & \textbf{RefCLIPScore} & \textbf{ClinicalBLEURT} & \textbf{MedBERTScore} \\
\hline
pclmed & 634 & 0.629913 & 0.272626 & 0.268994 & 0.337626 & 0.113264 & 0.268133 & 0.823614 & 0.817610 & 0.466557 & 0.632318 \\
\hline
CS\_Morgan & 429 & 0.628059 & 0.250801 & 0.209298 & 0.317385 & 0.092682 & 0.245029 & 0.821262 & 0.815534 & 0.455942 & 0.632664 \\
\hline
\textbf{DarkCow} & \textbf{220} & \textbf{0.626720} & \textbf{0.245228} & \textbf{0.195044} & \textbf{0.306005} & \textbf{0.088897} & \textbf{0.224250} & \textbf{0.818440} & \textbf{0.811700} & \textbf{0.456199} & \textbf{0.629189} \\
\hline
auebnlpgroup & 630 & 0.621112 & 0.204883 & 0.111034 & 0.289907 & 0.068022 & 0.176923 & 0.804067 & 0.798684 & 0.486560 & 0.626134 \\
\hline
2Q2T & 643 & 0.617814 & 0.247755 & 0.221252 & 0.313942 & 0.098590 & 0.220037 & 0.827074 & 0.813756 & 0.475908 & 0.622447 \\
\hline
MICLab & 678 & 0.612850 & 0.213525 & 0.185269 & 0.306743 & 0.077181 & 0.158239 & 0.815925 & 0.804924 & 0.445257 & 0.617195 \\
\hline
DLNU\_CCSE & 674 & 0.606578 & 0.217857 & 0.151179 & 0.283133 & 0.070419 & 0.168765 & 0.796707 & 0.790424 & 0.475625 & 0.612954 \\
\hline
Kaprov & 559 & 0.596362 & 0.190497 & 0.169726 & 0.295109 & 0.060896 & 0.107017 & 0.792183 & 0.787201 & 0.439971 & 0.608924 \\
\hline
DS@BioMed & 571 & 0.579438 & 0.103095 & 0.012144 & 0.220211 & 0.035335 & 0.071529 & 0.775566 & 0.774823 & 0.529529 & 0.580388 \\
\hline
DBS-HHU & 637 & 0.576891 & 0.153103 & 0.149275 & 0.270965 & 0.055929 & 0.064361 & 0.784199 & 0.774985 & 0.476634 & 0.588744 \\
\hline
KDE-medical-caption & 557 & 0.567329 & 0.132496 & 0.106025 & 0.256576 & 0.038628 & 0.038404 & 0.765059 & 0.760958 & 0.502234 & 0.569659 \\
\hline
\end{tabular}
\end{adjustbox}
\end{table}
This paper is organized explicitly as follows: Section~\ref{sec2} presents an overview of studies related to our research field. In Section~\ref{sec3}, we introduce the data process and some detailed analysis of our dataset. Next, Section~\ref{sec4} introduces some image pre-processing techniques. Section~\ref{sec5} details the design of the proposed methods and evaluation metric. Section~\ref{sec6} Present experimental results based on the proposed method. Section~\ref{sec7} discusses some impact. Finally, Section~\ref{sec8} summarizes the research and suggests future directions.
\section{Background and Related Works}\label{sec2}
\subsection{Radiology Techniques}

With the continuous advancement of imaging technology, medical imaging diagnosis has evolved from a supplementary examination tool to the most important clinical diagnostic and differential diagnostic method in modern medicine. Radiology techniques are used to scan images within the body, which are then interpreted and reported by radiologists to specialists \cite{kasban2015comparative}. With advancements in imaging technology, various imaging diagnostic methods have been developed, each with its own advantages and limitations. For example, X-ray imaging \cite{seibert2005x} offers non-invasive, quick, and painless imaging, but it involves exposure to ionizing radiation, which increases the risk of developing cancer later in life. On the other hand, MRI imaging \cite{atkins1998fully} provides non-ionizing radiation and high spatial resolution, but it has relatively low sensitivity and longer scanning times, etc.

\subsection{Former Medical Image Captioning Datasets}

Medical imaging diagnosis today plays an incredibly important role in both the healthcare and information technology sectors. It not only aids in diagnosis and increases understanding of diseases but also holds immense potential in improving healthcare delivery and enhancing quality of life. The application of deep learning in medical image captioning in an era where AI is ubiquitous is evident; it automates the annotation process and significantly accelerates image analysis. Several datasets have been created to facilitate the training of medical image captioning tasks such as ROCO \cite{pelka2018radiology}, PadChest \cite{bustos2020padchest}, MIMIC-CXR \cite{johnson2019mimic}, IU X-Ray \cite{wijerathna2022chest}, and MedICaT \cite{subramanian2020medicat}.

\subsection{Related Work Methods}
For the task of medical image captioning, various methods have been developed, with pioneering work in applying the CNN-RNN encoder-decoder approach to generate captions from medical images conducted by \citet{shin2016learning}. They utilized either the Network-in-Network or GoogLeNet architectures as encoding models, followed by LSTM \cite{graves2012long} or GRU \cite{dey2017gate} as the decoding RNN to translate the encoded images into descriptive captions. In the process of translating images into biomedical text, MDNET made a notable advancement by incorporating an attention mechanism called MDNET \cite{zhang2017tandemnet}. This model employs RESNET for image encoding, extending its skip connections to mitigate gradient vanishing.

In recent studies by \citet{Wang2021ImageSemGA}, \citet{kougia-2019-aueb}, and \citet{NEURIPS2018_e0741335}, a fusion of generative models and retrieval systems for Medical Image Captioning (MIC) has been explored. For instance, \citet{Wang2021ImageSemGA} proposed an approach that alternates between template retrieval and sentence generation for rare abnormal descriptions. This method relies on a contextual relational-topic encoder derived from visual and textual features, facilitating semantic consistency through hybrid knowledge co-reasoning. Additionally, \citet{kougia-2019-aueb} from AUEB NLP group presented various systems for the Image-CLEFmed 2019 Caption task. One approach utilized a retrieval-based model that leverages visual features to retrieve the most similar images based on cosine similarity, combining their concepts to predict relevant captions. Another system incorporated CheXNet with enhanced classification labels, employing a CNN encoder and a feed-forward neural network (FFNN) for multi-label classification. They also suggested an ensemble model by combining these systems, computing scores for returned concepts and merging them with image similarity scores to select the most relevant concepts.

Large language models (LLMs) have catalyzed significant progress in medical question answering; Med-PaLM \citep{singhal2023large} was the first model to exceed a “passing” score in US Medical Licensing Examination (USMLE). However, this and other prior work suggested significant room for improvement, especially when models’ answers were compared to clinicians’ answers. Med-PaLM 2 \citep{singhal2023towards} bridges these gaps by leveraging a combination of base LLM improvements, medical domain finetuning, and prompting strategies including a novel ensemble refinement approach. Base LLM For Med-PaLM \citep{singhal2023large}, the base LLM was PaLM, Med-PaLM 2 \citep{singhal2023towards} builds upon PaLM 2, a new iteration of Google’s large language model with substantial performance improvements on multiple LLM benchmark tasks.

\section{Dataset}\label{sec3}
Thanks to AUEB NLP Group for providing an excellent analysis of the dataset in the study of  \citet{kaliosis2023aueb}. When comparing ImageCLEFmedical2023 data with ImageCLEFmedical2024, we found no significant differences in the task of caption prediction. Therefore, we decided to reapply to analyze the dataset in this section.

\begin{figure}[htp]
    \centering
    \includegraphics[width=1\linewidth]{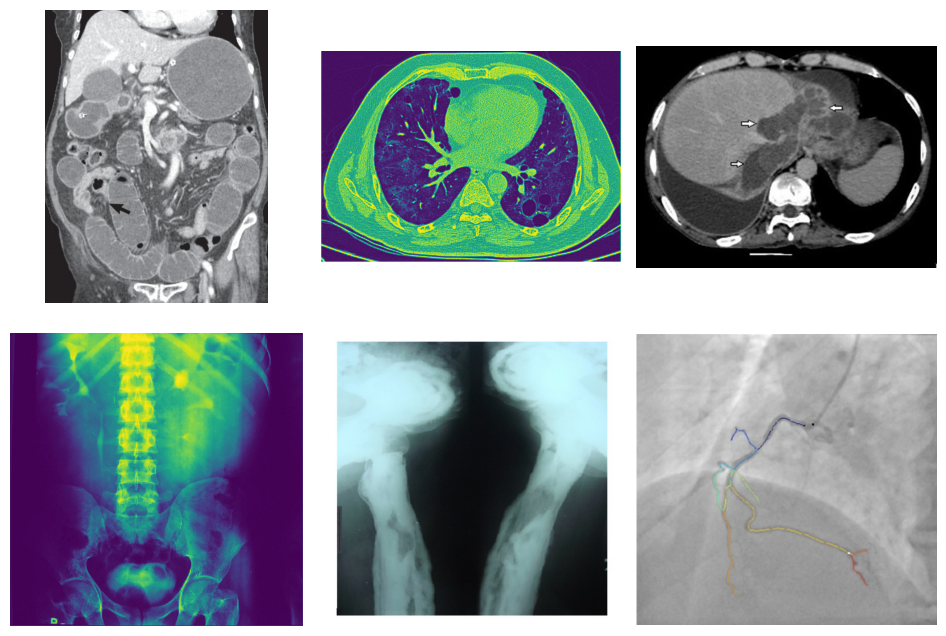}
    \caption{Several images from the ImageCLEFmedical2024 dataset}
    \label{example}
\end{figure}

This year's ImageCLEFmedical Caption task provided a dataset that includes 70,108 radiology images in the training set, each annotated with medical concepts using UMLS terms and diagnostic captions. The organizers initially divided the dataset into training and validation subsets. Building on previous campaigns, this year's dataset is an updated and expanded version of the Radiology Objects in Context (ROCO) dataset, which is sourced from a variety of biomedical studies in the PubMed Central OpenAccess (PMC OA) subset. The dataset used for the caption prediction task includes images from different modalities, such as X-ray and Computed Tomography (CT), although specific details about the image types were not provided. The goal of the caption prediction task is to generate open-ended diagnostic texts for the medical images (see Figure \ref{example}).

\begin{table}[htp]
\centering
\caption{The ten most common words and their frequencies in the ImageCLEFmedical2024 train set.}
\label{tab:most_common_words}
\begin{adjustbox}{width=1\textwidth}
\begin{tabular}{|l|*{10}{c|}}
\hline
\multicolumn{11}{|c|}{\textbf{Most common words (excluding stop-words)}} \\ \hline
\textbf{Word} &
  \textbf{showing} &
  \textbf{right} &
  \textbf{left} &
  \textbf{ct} &
  \textbf{image} &
  \textbf{chest} &
  \textbf{scan} &
  \textbf{computed} &
  \textbf{tomography} &
  \textbf{shows} \\ \hline
\textbf{Occurrences} &
  22519 &
  18258 &
  18136 &
  15167 &
  10245 &
  10082 &
  9296 &
  9273 &
  8969 &
  8600 \\ \hline
\end{tabular}
\end{adjustbox}
\end{table}

\begin{table}[htp]
\centering
\caption{The five most common captions found in the ImageCLEFmedical2024 train set alongside the number of images they are associated with.}
\label{five most common captions}
\begin{adjustbox}{width=0.6\textwidth}
\begin{tabular}{|clc|}
\hline
\multicolumn{3}{|c|}{\textbf{Most common captions}}                                       \\ \hline
\multicolumn{1}{|c|}{\textbf{Position}} & \multicolumn{1}{c}{\textbf{Caption}} & \textbf{Occurences} \\ \hline
\multicolumn{1}{|c|}{\textbf{1}} & \multicolumn{1}{l|}{Initial panoramic radiograph} & 40 \\
\multicolumn{1}{|c|}{\textbf{2}} & \multicolumn{1}{l|}{Final panoramic radiograph}   & 37 \\
\multicolumn{1}{|c|}{\textbf{3}} & \multicolumn{1}{l|}{Chest X-ray}                  & 20 \\
\multicolumn{1}{|c|}{\textbf{4}} & \multicolumn{1}{l|}{Chest radiograph}             & 17 \\
\multicolumn{1}{|c|}{\textbf{5}} & \multicolumn{1}{l|}{Preoperative CT scan.}                  & 9  \\ \hline
\end{tabular}%
\end{adjustbox}
\end{table}

In the Caption prediction sub-task, each image has a diagnostic caption describing the described medical condition. There are a total of 69,743 captions in the training dataset and 9,959 captions in the validation dataset, one for each image. Similar to last year's campaign, the majority of captions (99.47\%, or 69,743 out of 70,108) were unique. This is a notable difference from previous versions of the quest, where the uniqueness percentage was much lower. As a result, traditional retrieval methods based on nearest neighbor search are less efficient this year, including variants with a weighting mechanism based on the cosine similarity of the retrieved images. Therefore, more complex methods of creating subtitles are needed.
\begin{figure}[htp]
    \centering
    \includegraphics[width=1\linewidth]{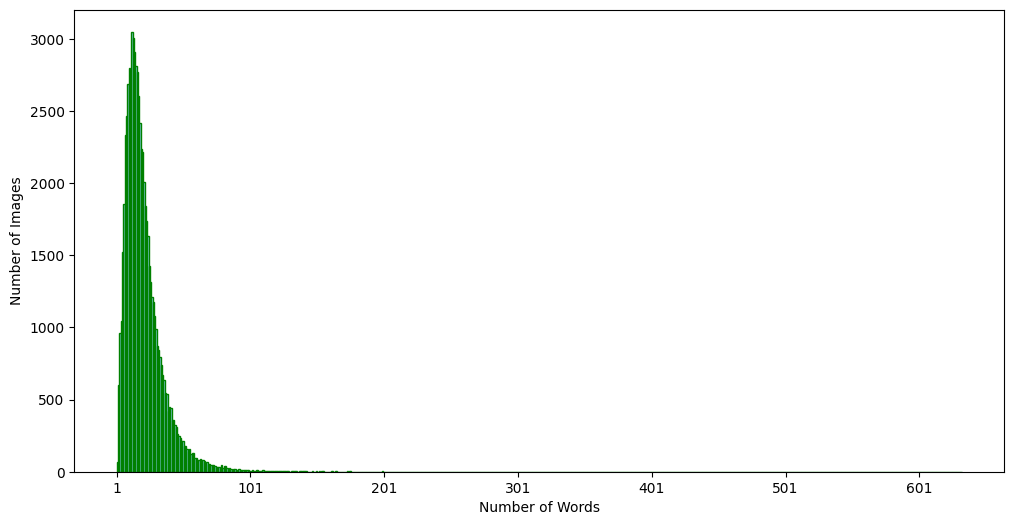}
    \caption{Distribution of caption lengths in the training set}
    \label{Distribution of caption lengths in the training set}
\end{figure}

\begin{figure}[htp]
    \centering
    \includegraphics[width=1\linewidth]{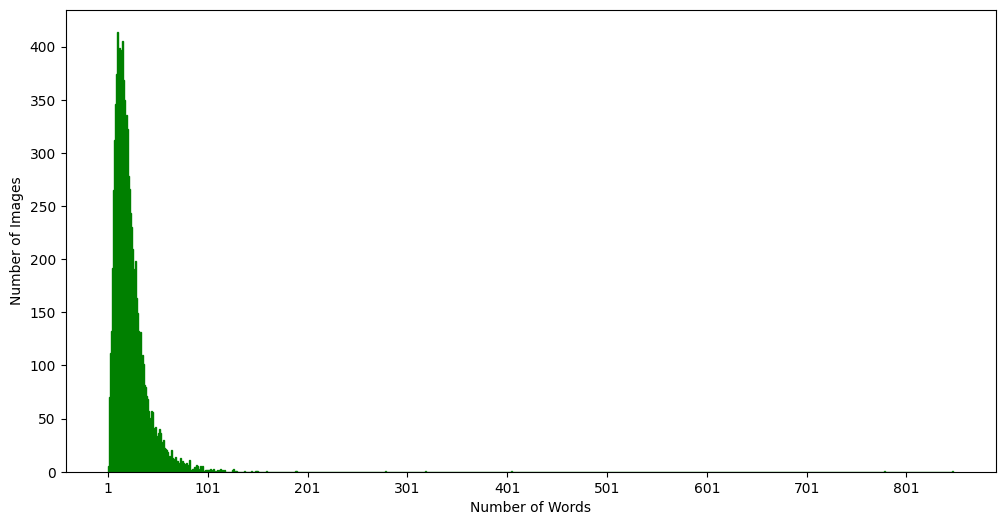}
    \caption{Distribution of caption lengths in the valid set}
    \label{Distribution of caption lengths in the valid set}
\end{figure}

We observed that the maximum number of words in a single caption is 848 (occurred once), while the minimum is 1 (encountered 1 time). The average caption length is 20.84 words. These statistics apply to the entire dataset ( training set and valid set). The five most common captions, as well as the ten most popular words (excluding stopwords), can be found in Tables \ref{five most common captions} and \ref{tab:most_common_words}, respectively. In Figure \ref{Distribution of caption lengths in the training set} and Figure \ref{Distribution of caption lengths in the valid set}, we present a distribution caption length of the training and valid sets, both indicating that the majority of captions contain fewer than 100 words.
\section{Image Pre-processing} \label{sec4}
\subsection{Denoising}
\begin{figure}[htp]
  \centering
  \includegraphics[width=1.0\textwidth]{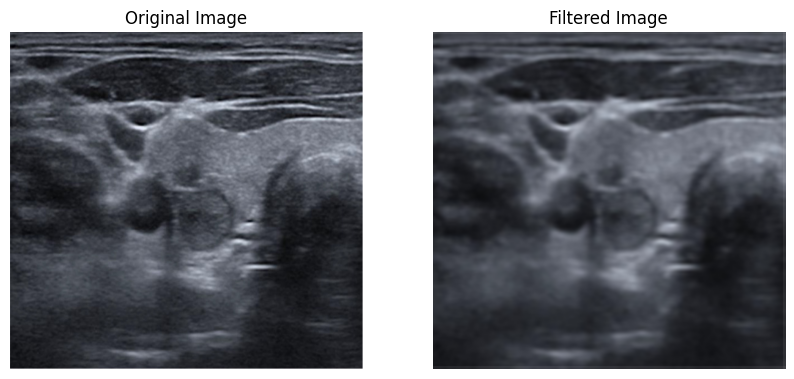}
  \caption{Application of Gaussian filter.}
  \label{gauss_filter}
\end{figure}
Denoising is crucial in enhancing image quality by reducing the noise while preserving the important details. Noise in medical images can come from various sources, such as sensor imperfections, poor scan conditions, or inherent patient movements during image acquisition.

The smoothness of images is controlled through the utilization of a Gaussian filter with a fixed kernel size. The Gaussian filter operates by smoothing images using a technique called convolution. It employs a Gaussian kernel - a matrix based on the Gaussian function to adjust pixel values. This kernel is applied over each pixel in the image, averaging the pixel values in its vicinity, weighted by their distance from the central pixel. The standard deviation $\sigma$ of the Gaussian determines the amount of blurring: a larger $\sigma$ results in more blurring, smoothing out more details and noise. This process helps in reducing noise and is often used as a preparatory step in image processing tasks to enhance image quality without losing critical structural details (see Figure \ref{gauss_filter}).

The 2-D Gaussian function is given by:
\begin{equation}
G(x,y) = \frac{1}{2\pi\sigma^2} e^{-\frac{x^2+y^2}{2\sigma^2}}
\end{equation}

\begin{figure}[htp]
  \centering
  \includegraphics[width=1.0\textwidth]{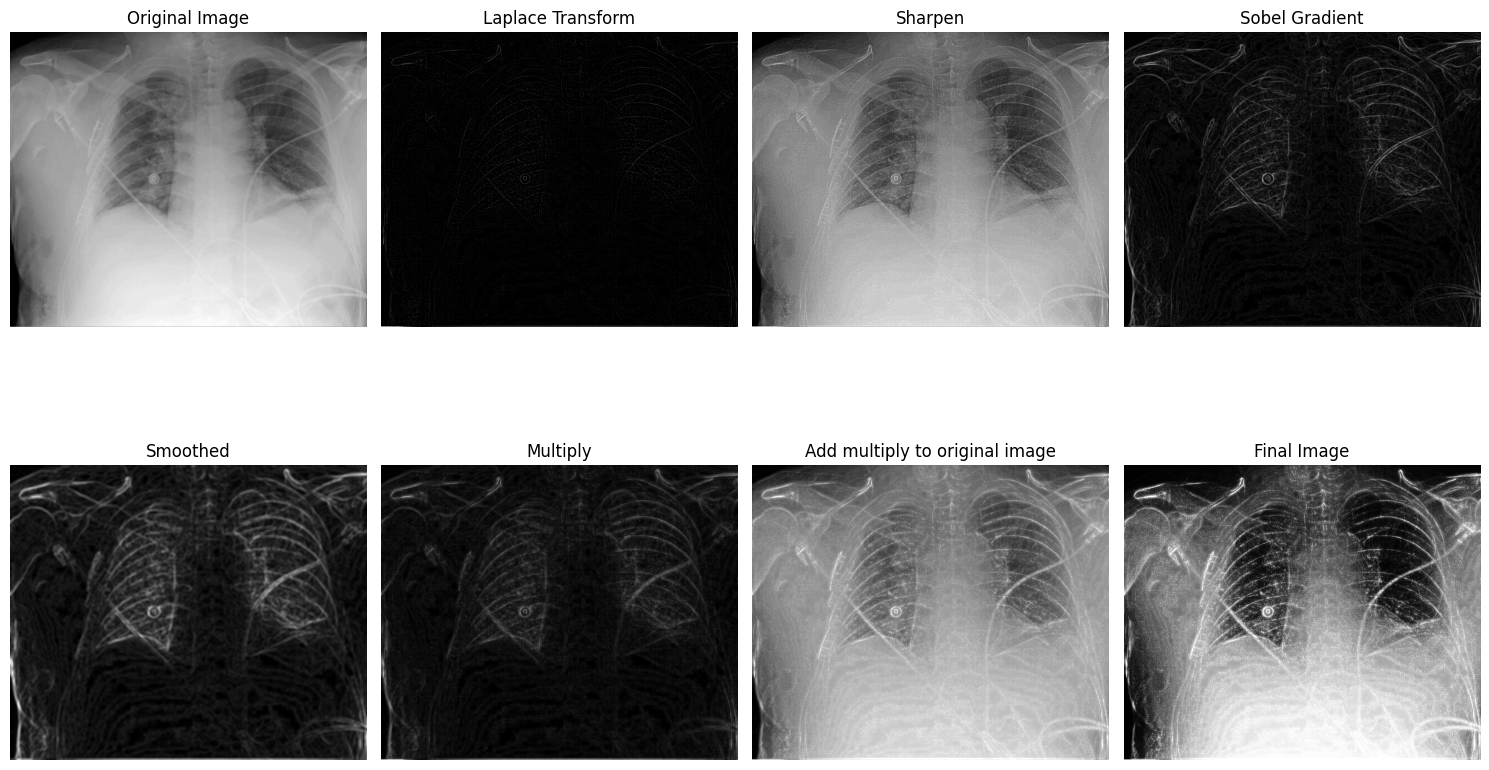}
  \caption{The image after a series of processing.}
  \label{image_enhancement}
\end{figure}

Medical image enhancement is one of the most widely used medical image processing techniques in medical domain. Its purpose is to improve the visual effect of the image and facilitate the analysis and understanding of the image by humans or machines. The Laplace transform and the Sobel gradient operator are two common ways of performing edge detection, image sharpening, and enabling the enhancement of the image (see Figure \ref{image_enhancement}). 

\textbf{Step 1 Laplace Transform:} Apply the Laplace transform to enhance contrast by emphasizing areas of rapid intensity change in the original image.

\textbf{Step 2 Sobel Operator:}
Use the Sobel operator to enhance the edges of the image. This step also helps to smooth out noise, making the edges clearer and more cohesive.

\textbf{Step 3 Smoothing:}
Smooth the image processed by the Sobel operator using a 3x3 mean filter. This step increases the contrast of the edges against the background.

\textbf{Step 4 Dot Product:}
Intensify the contrast by performing a dot product of the smoothed image with the result from the Laplace transform from step 1.

\textbf{Step 5 Addition for Final Sharpening:} Enhance the sharpness and visibility of detail by adding the result of the dot product back to the original image.

\textbf{Step 6 Histogram Equalization:}
Apply histogram equalization to distribute the histogram of the image uniformly, improving the overall contrast and making fine details more visible.

\subsection{Image Enhancement}

\section{Proposed Method} \label{sec5}
\subsection{Encoder-Decoder Approach}
\subsubsection{Features Embedding}
\begin{figure*}[htp]
  \centering
  \includegraphics[width=1.0\textwidth]{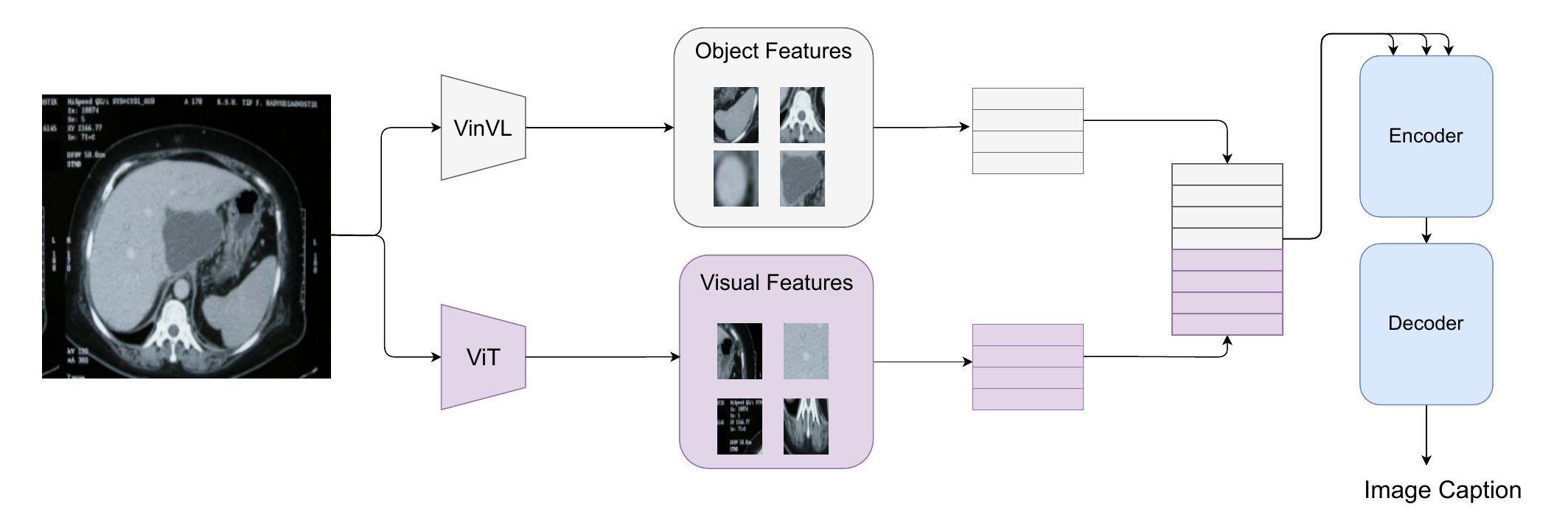}
  \caption{Overview of VisionDiagnostor.}
  \label{VisionCaptioner structure}
\end{figure*}

We propose VisionDiagnostor centers around the implementation of Transformer encoder-decoder approach and deployed to evaluate methods having \textbf{ClinicalT5} \cite{lehman2023clinical} and \textbf{BioBART} \cite{yuan2022biobart} as encoder-decoder module (see Figure \ref{VisionCaptioner structure}). \textbf{ClinicalT5}, based on the \textbf{T5} \cite{raffel2020exploring} architecture, and \textbf{BioBART}, a variant of the \textbf{BART} \cite{lewis2019bart} architecture, have both been trained on a large of biomedical text data. These models stand out as the preeminent and potent pre-trained language models for the medical domain, ensuring the efficacy and robustness of our proposed method.

\textbf{Object features: }To extract object features in an image, we used the VinVL model to extract object features \(R = \{r_1, r_2, ..., r_k\}\) from an image, with each \(r_i\) being a 2048-dimensional vector. Bounding box coordinates are normalized as \(b_i = \left[\frac{{x_i^{min}}}{w}, \frac{{y_i^{min}}}{h}, \frac{{x_i^{max}}}{w}, \frac{{y_i^{max}}}{h}\right]\), forming \(B_{\text{obj}} = \{b_1, b_2, ..., b_k\}\).

Final object features \(V_{\text{obj}}\) are computed by projecting \(R\) and \(B_{\text{obj}}\) to the language model dimension and summing the results:

\begin{align}
V_{\text{obj}} = R' + B'_{\text{obj}}
\end{align}

We use ViT for visual feature extraction due to its ability to capture global information through its attention mechanism. By freezing ViT and projecting the last hidden state to match the language model's dimension, we obtain visual features \(V\).

The input embedding to the encoder-decoder module is:

\begin{align}
& \text{Input} = \text{Concat}(V, V_{\text{obj}})
\end{align}

Where \(V\) are the visual features from ViT, and \(V_{\text{obj}}\) are the VinVL region object features. The \(\text{Concat}(\cdot)\) function concatenates these features.

\subsubsection{Encoder-Decoder Module}
In this task, we employed the Transformer encoder-decoder architecture, which is used in ClinicalT5 \cite{lehman2023clinical} and BioBART \cite{yuan2022biobart} for the encoder-decoder module of VisionDiagnostor. The encoder receives the input features and then passes them to the decoder to generate the output sentence. In the decoder, attention mechanisms are employed, directing focus to both the output of the encoder and the input of the decoder.

\subsection*{\textbf{Encoder}}
\subsection*{\textbf{Multi-Head Attention:}}
\begin{align}
& \text{Attention}^{(\text{Enc})}(Q,K,V) = \text{softmax}\left(\frac{QK^T}{\sqrt{d_k}}\right)V
\end{align}
where \( Q \), \( K \), and \( V \) are the query, key, and value matrices, and \( d_k \) is the dimensionality of the key vectors.

\subsection*{\textbf{Encoder Feed-Forward Network:}}
\begin{align}
& \text{FFN}^{(\text{Enc})}(x) = \text{ReLU}(xW_1^{(\text{Enc})} + b_1^{(\text{Enc})})W_2^{(\text{Enc})} + b_2^{(\text{Enc})}
\end{align}

where \( W_1^{(\text{Enc})} \), \( W_2^{(\text{Enc})} \), \( b_1^{(\text{Enc})} \), and \( b_2^{(\text{Enc})} \) are learnable parameters.

\subsection*{\textbf{Encoder Layer Normalization:}}
\begin{align}
& \text{LayerNorm}^{(\text{Enc})}(x) = \text{LN}^{(\text{Enc})}(x + \text{LayerNorm}^{(\text{Enc})}(x))
\end{align}

where \( \text{LN}^{(\text{Enc})} \) is the layer normalization function.

\subsection*{\textbf{Decoder}}
\subsection*{\textbf{Decoder Self-Attention:}}
\begin{align}
\text{Attention}^{(\text{Dec})}(Q,K,V) = \text{softmax}\left(\frac{QK^T}{\sqrt{d_k}}\right)V
\end{align}
where \( Q \), \( K \), and \( V \) are the query, key, and value matrices, and \( d_k \) is the dimensionality of the key vectors.

\subsection*{\textbf{Decoder-Encoder Cross-Attention:}}
\begin{align}
& \text{Attention}^{(\text{Dec})}(Q,K,V) = \text{softmax}\left(\frac{QK^T}{\sqrt{d_k}}\right)V 
\end{align}

where \( Q \) comes from the decoder and \( K \), \( V \) come from the encoder.

\subsection*{\textbf{Decoder Feed-Forward Network:}}
\begin{align}
& \text{FFN}^{(\text{Dec})}(x) = \text{ReLU}(xW_1^{(\text{Dec})} + b_1^{(\text{Dec})})W_2^{(\text{Dec})} + b_2^{(\text{Dec})}
\end{align}

where \( W_1^{(\text{Dec})} \), \( W_2^{(\text{Dec})} \), \( b_1^{(\text{Dec})} \), and \( b_2^{(\text{Dec})} \) are learnable parameters.

\subsection*{\textbf{Decoder Layer Normalization:}}
\begin{align}
\text{LayerNorm}^{(\text{Dec})}(x) = \text{LN}^{(\text{Dec})}(x + \text{LayerNorm}^{(\text{Dec})}(x))
\end{align}
where \( \text{LN}^{(\text{Dec})} \) is the layer normalization function.

\subsection{Query Transformer Approach}
Inspired by the BLIP2 architecture \cite{li2023blip}, we leveraged the Query Transformer (Q-Former) module, which serves as the trainable intermediary between a fixed image encoder and a fixed Large Language Model. It extracts a consistent number of output features from the image encoder, irrespective of the input image resolution. Q-Former comprises two transformer submodules that share self-attention layers: an image transformer for visual feature extraction from the fixed image encoder and a text transformer acting as both an encoder and decoder.
\begin{figure*}[htp]
  \centering
  \includegraphics[width=1.0\textwidth]{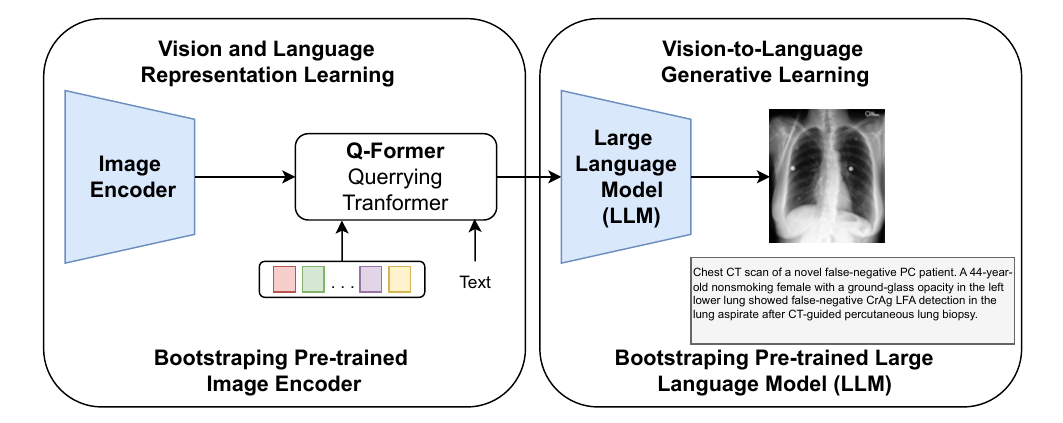}
  \caption{Overview of VisionDiagnostor-Q.}
  \label{BLIP2}
\end{figure*}

We initialize a set number of learnable query embeddings as input to the image transformer. These queries engage in self-attention and cross-attention interactions with each other and the frozen image features. Additionally, they can interact with the text through self-attention layers, with different attention masks applied based on the pre-training task.

In our experiments, we employ 64 queries, each with a dimensionality of 768, matching the hidden dimension of Q-Former. We utilize VIT-huge \cite{he2022masked} as the frozen image encoder and BioMistral-7B \cite{labrak2024biomistral} as the frozen LLM for caption generation, and we call it VisionDiagnostor-Q-BioMistral which is depicted in Figure \ref{BLIP2}. This bottleneck architecture, combined with our pre-training objectives, compels the queries to extract visual information most pertinent to the accompanying text.

\subsection{Evaluation Metrics}
\subsubsection{BERTScore}
BERTScore is computed as proposed by  \citet{zhang2019BERTScore}, where the cosine similarity of each hypothesis token $j$ with each token $i$ in the reference sentence is calculated using contextualized embeddings. Instead of using a time-consuming best-case matching approach, a greedy matching strategy is employed. The F1 measure is then calculated as follows:

\begin{align}
& R_{\text{BERT}} = \frac{1}{|\mathbf{r}|} \sum_{i \in \mathbf{r}} \max_{j \in \mathbf{p}} \cos(\vec{i}, \vec{j}), \\
& P_{\text{BERT}} = \frac{1}{|\mathbf{p}|} \sum_{j \in \mathbf{p}} \max_{i \in \mathbf{r}} \cos(\vec{i}, \vec{j}), \\
& \text{BERTScore} = F_{\text{BERT}} = \frac{2 \cdot P_{\text{BERT}} \cdot R_{\text{BERT}}}{P_{\text{BERT}} + R_{\text{BERT}}}.
\end{align}

The BERTScore correlates better with human judgments for the tasks of image captioning and machine translation.

\subsubsection{Other Metrics}
In addition to BERTScore, the competition also uses many other metrics such as ROUGE \cite{lin2004rouge}, BLEU-1 \cite{papineni2002bleu}, BLEURT \cite{sellam2020bleurt}, METEOR \cite{banerjee2005meteor}, CIDEr \cite{vedantam2015cider}, CLIPScore \cite{hessel2021clipscore}, RefCLIPScore \cite{hessel2021clipscore}, ClinicBLEURT \cite{huang2019clinicalbert} and MedBERTScore \cite{abacha2023investigation}. Applying a variety of these metrics helps us have a more accurate and general view of the model performance of participating teams. Each measure has its own advantages and provides a different perspective on text quality that makes it relevant in a medical context. This multi-dimensional evaluation helps identify outstanding models and gain an objective view of the competition.
\section{Experiment Results} \label{sec6}
\subsection{Experimental Configuration}
All our proposed methods were trained and fine-tuned using the Adam optimization \cite{kingma2014adam}. We utilized an A100-GPU setup with 80GB of memory to train models, taking 10 hours on average for each method. We set the learning rate to 3e-05, dropout is set at 0.2, batch size is 32, and the training process is terminated after 3 epochs of not finding any reduction in the valid loss.
\subsection{Main Result}
\begin{table}[htp]
\centering
\caption{Performance comparison of different models on test set, VD stands for VisionDiagnostor.}
\label{main_results}
\begin{adjustbox}{width=1\textwidth}
\begin{tabular}{lccccccccccc}
\hline
Model&Model Size & BERTScore & ROUGE & BLEU-1 & BLEURT & METEOR & CIDEr & CLIPScore & RefCLIPScore & ClinicalBLEURT & MedBERTScore \\ \hline
VD-Q-BioMistral&8B & 0.6200 & 0.2139 & 0.1685 & 0.2913 & 0.0751 & 0.1585 & 0.8132 & 0.8014 & \textbf{0.4597} & 0.6233 \\ 
VD-ClinicalT5&310M & 0.5994 & 0.2363 & \textbf{0.2323} & 0.2954 & \textbf{0.0989} & 0.1442 & \textbf{0.8244} & 0.8100 & \textbf{0.4597} & 0.6016 \\ 
VD-BioBART&227M & \textbf{0.6267} & \textbf{0.2452} & 0.1950 & \textbf{0.3060} & 0.0889 & \textbf{0.2243} & 0.8184 & \textbf{0.8117} & 0.4562 & \textbf{0.6292} \\ 
 \hline
\end{tabular}
\end{adjustbox}
\end{table}

Table \ref{main_results} presents a comprehensive of the results achieved by individual models, showcasing their BERTScore and other metrics. The findings underscore significant disparities in performance among the various models, providing valuable insights into their respective strengths and weaknesses. Notably, within the baseline models, VisionDiagnostor-BioBART stands out as the top performer, showcasing an impressive BERTScore of 0.6267 and almost all other metrics with the smallest size at 227M parameters. Moreover, Table \ref{main_results} demonstrates that using large-scale pre-trained models in VisionDiagnostor-Q-BioMistral with a very large size (8B) does not result in significant performance improvement in this task. 
\section{Result Analysis} \label{sec7}
In this section, we conduct a subjective analysis of the valid set due to the limited number of submissions in the competition. This means that instead of using the test set for objective evaluation, we used the valid set to analyze the results our proposed methods achieved.
\subsection{Impact of Image Pre-processing}
\begin{table}[htp]
\centering
\caption{Results of models with image pre-processing in valid set. $\triangle$ indicates the increase (↑) or decrease (↓) and compares without pre-processing (*).}
\label{Table:bang5}
\begin{tabular}{lcc}
\hline
\multirow{2}{*}{\textbf{Model}} & \multirow{2}{*}{\textbf{BERTScore}} \\ 
 & \textbf{} \\ \hline
\begin{tabular}[c]{@{}l@{}}VisionDiagnostor-Q-BioMistral\\ $\triangle$\end{tabular} &
  \begin{tabular}[c]{@{}c@{}}0.6841*\\ \textcolor{red}{$\downarrow 0.0101$}\end{tabular} \\
\begin{tabular}[c]{@{}l@{}}VisionDiagnostor-ClinicalT5 \\ $\triangle$\end{tabular} &
  \begin{tabular}[c]{@{}c@{}}0.7071*\\ \textcolor{red}{$\downarrow 0.0166$}\end{tabular} \\
\begin{tabular}[c]{@{}l@{}}VisionDiagnostor-BioBART\\ $\triangle$\end{tabular} &
  \begin{tabular}[c]{@{}c@{}}0.7165*\\ \textcolor{blue}{$\uparrow 0.0198$}\end{tabular} \\
\hline
\end{tabular}
\end{table}

Table \ref{Table:bang5} presents the results comparing the performance of the models with and without image pre-processing on the validation dataset, evaluated using BERTScore. Specifically, for the VisionDiagnostor-Q-BioMistral model, BERTScore decreased from 0.6841 to 0.6740 after applying pre-processing, corresponding to a decrease of 0.0101. Similarly, VisionDiagnostor-ClinicalT5 also saw a decrease in performance from 0.7071 to 0.6905, a decrease of 0.0166. In contrast, VisionDiagnostor-BioBART is the only model with an improvement with BERTScore increasing from 0.7165 to 0.7363, an increase of 0.0198.

Overall, applying image pre-processing does not appear to yield significant improvement for most models. Even for two of the three models (VisionDiagnostor-Q-BioMistral and VisionDiagnostor-ClinicalT5), image pre-processing degrades performance. The reason may be because of the input images are of good quality and have almost no noise. Some images also have clear instructions, such as arrows pointing to the relevant caption of the image (see Figure \ref{example} in Section \ref{sec3}), making it easy for the model to understand and process the content without additional pre-processing.

\subsection{Impact of Caption Length}
\begin{table}[htp]
  \centering
  \caption{Group of caption length in valid set.}
  \begin{tabular}{lrrr}
    \hline
    \textbf{Group} & \textbf{Length (n)} & \textbf{Samples}\\
    \hline
    Short & $n \leq 20$ & 5520 \\
    Medium & $20 < n \leq 25$ & 2179 \\
    Long & $25 < n \leq 30$ &1339 \\
    Very long & $n > 30$& 934 \\
    \hline
  \end{tabular}
  \label{cap_len}
\end{table}

The details of the test set based on different groups of lengths are in Table \ref{cap_len}. Classification is done as follows:

\begin{itemize}
\item Short caption: These are captions shorter than 21 words.

\item Medium caption: This group includes captions from 21 to 25 words.

\item Long caption: Captions in this group from 26 to 30 words. 

\item Very long caption: This group contains captions longer than 30 words.
\end{itemize}

\begin{figure}[htp]
  \centering
  \includegraphics[width=0.7\textwidth]{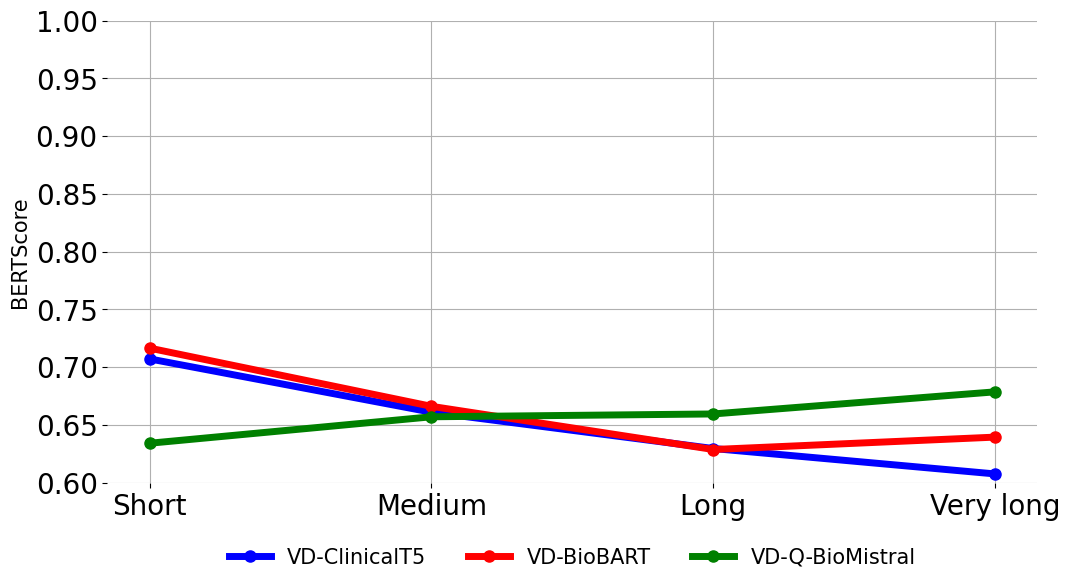}
  \caption{The results of models based on caption length.}
  \label{caption_len}
\end{figure}

The illustration from Figure \ref{caption_len} is an important step in gaining insight into the model's performance for different caption lengths. The results show that the length of the caption plays an important role in influencing model performance.

Specifically, the two models VisionDiagnostor-ClinicalT5 and VisionDiagnostor-BioBART based on the encoder-decoder method have similar trends, both showing a gradual decrease in BERTScore score as the caption length increases. This may indicate a limitation in handling longer captions with this method.

It is worth noting that the VisionDiagnostor-Q-BioMistral model represents a different case, with performance increasing as the caption length increases. This may imply that this model is capable of handling longer captions more efficiently than other models, possibly due to its complexity and magnitude.

\subsection{Impact of Object Features}
According to papers from competing teams in previous years \cite{nicolson2023concise}, \cite{zhou2023transferring}, \cite{kaliosis2023aueb}, the most popular image feature extraction methods today have two main directions: convolutional neural networks (CNN) and Vision transformers (ViT). Studies and demonstrations have shown that ViT often gives better results than CNN in the task of image captioning. ViT is capable of capturing long-term and global relationships in images more effectively, leading to the creation of richer and more accurate captions. However, to improve the quality of feature extraction further, we used the VinVL model. VinVL takes advantage of the power of the ability to detect and represent objects in images in detail. This allows the model to gain a deeper understanding of the context and elements in the image, thereby creating more accurate captions.
\begin{table}[htp]
\centering
\caption{Results of models with image pre-processing in valid set. $\triangle$ indicates the increase (↑) or decrease (↓) and compares with models using object features (*).}
\label{Table:bang7}
\begin{tabular}{lcc}
\hline
\multirow{2}{*}{\textbf{Model}} & \multirow{2}{*}{\textbf{BERTScore}} \\ 
 & \textbf{} \\ \hline
\begin{tabular}[c]{@{}l@{}}VisionDiagnostor-ClinicalT5 \\ $\triangle$\end{tabular} &
  \begin{tabular}[c]{@{}c@{}}0.7071*\\ \textcolor{red}{$\downarrow 0.0239$}\end{tabular} \\
\begin{tabular}[c]{@{}l@{}}VisionDiagnostor-BioBART\\ $\triangle$\end{tabular} &
  \begin{tabular}[c]{@{}c@{}}0.7165*\\ \textcolor{red}{$\downarrow 0.0321$}\end{tabular} \\
\hline
\end{tabular}
\end{table}

Table \ref{Table:bang7} presents the results of the models when not using object features in the valid set. The figures show that not using object features significantly reduced the performance of the models.

Specifically, the VisionDiagnostor-ClinicalT5 model has a BERTScore of 0.7071 when using object features. However, when not using object features, the performance of this model drops by 0.0239. Similarly, the VisionDiagnostor-BioBART model also shows a significant decrease when not using object features, with BERTScore decreasing from 0.7165 to 0.6844, corresponding to a decrease of 0.0321.

These results indicate that using object features has an important effect in improving model performance. Object features can provide detailed and characteristic information about objects in images, helping models understand and describe images more accurately. Removing object features results in the loss of important information, reducing the model's ability to produce accurate and detailed captions, which in turn reduces BERTScore significantly.

\section{Conclusion and Future Works} \label{sec8}
In this paper, we have proposed three different models to solve the task of medical image captioning, in other words medical image diagnosis, including VisionDiagnostor-ClinicalT5 and VisionDiagnostor-BioBART based on encoder-decoder architecture, VisionDiagnostor-Q-BioMistral based on BLIP2 architecture with Query Transformer which leveraging the power of Large Language Models (LLM).

Our results show that the VisionDiagnostor-BioBART model achieved third place on the leaderboard, with the highest BERTScore of 0.6267, despite being the smallest in size with only 227M parameters. Additionally, we performed analysis of the results to gain a deeper understanding of the factors that influence the performance of the models, including image pre-processing, caption length, and object features. These analyses have provided the comprehensive insight needed to shape and improve future methods and models for this task.

In future works, our objective is to delve deeper into the applications of other biomedical large language models (LLMs) BioMedLM \cite{bolton2024biomedlm}, BioGPT \cite{luo2022biogpt}, especially focusing on enhancing their capabilities to generate precise captions that are context-sensitive. This development will be pursued through methods like instruction tuning and better alignment of the models with specific user requirements. In addition, we plan to explore the integration of dense retrieval techniques into the biomedical image captioning process \cite{moschovis2022neuraldynamicslab}. By adopting frameworks akin to Retrieval Augmented Generation, we intend to supplement the LLMs with an external, non-parametric memory using a FAISS index \cite{johnson2019billion}, thereby enriching their reasoning capabilities. Another area of interest will be investigating the interconnections between these approaches. We also anticipate evaluating the qualitative variations in the captions generated through these different methodologies to ascertain their efficacy and practicality in real-world applications.

\section*{Acknowledgment}\label{secA1}
This research was supported by The VNUHCM-University of Information Technology's Scientific Research Support Fund.

\bibliography{sample-ceur}

\end{document}